\pdfoutput=1

\documentclass[11pt]{article}

\usepackage[]{acl}

\usepackage{times}
\usepackage{latexsym}

\usepackage[T1]{fontenc}

\usepackage[utf8]{inputenc}

\usepackage{microtype}
\usepackage{graphicx}
\usepackage{subfigure}

\usepackage{amsmath,amsfonts,amssymb,bm}
\usepackage{pifont} 
\usepackage{graphicx,epsfig,fancybox} 
\usepackage{float}
\usepackage{color} 
\usepackage{multirow}
\usepackage{natbib}

\usepackage{tikz}
\usetikzlibrary{shapes.geometric, positioning, calc}
\usetikzlibrary{arrows,shapes,calc}
\usetikzlibrary{trees,positioning,mindmap,shadows,fit}
\usetikzlibrary{decorations.pathreplacing}
\usetikzlibrary{tikzmark} 
\usetikzlibrary{intersections} 

\newcommand{\revised}[1]{\textcolor{blue}{}}
\renewcommand{\vec}[1]{\boldsymbol{#1}}

\usepackage{adjustbox}
\usepackage{array}
\usepackage{booktabs}

\newcolumntype{R}[2]{%
    >{\adjustbox{angle=#1,lap=\width-(#2)}\bgroup}%
    l%
    <{\egroup}%
}

\usepackage{setspace}

\usepackage{trackchanges}
\addeditor{YJ}
\addeditor{HC}
\addeditor{RT}

\title{Identifying the Source of Vulnerability in Explanation Discrepancy: A Case Study in Neural Text Classification}

\author{Ruixuan Tang \\
  University of Virginia\\
  \texttt{rt5tb@virginia.edu} \\\And
  Hanjie Chen\\
  University of Virginia\\
  \texttt{hc9mx@virginia.edu} \And
  Yangfeng Ji\\
  University of Virginia\\
  \texttt{yangfeng@virginia.edu} \\}

\begin{document}
\maketitle
\begin{abstract}
Some recent works observed the instability of post-hoc explanations when input side perturbations are applied to the model. 
This raises the interest and concern in the stability of post-hoc explanations. 
However, the remaining question is: is the instability caused by the neural network model or the post-hoc explanation method?
This work explores the potential source that leads to unstable post-hoc explanations.  
To separate the influence from the model, we propose a simple \emph{output probability perturbation} method. 
Compared to prior input side perturbation methods, the \emph{output probability perturbation} method can circumvent the neural model's potential effect on the explanations and allow the analysis on the explanation method. 
We evaluate the proposed method with three widely-used post-hoc explanation methods (LIME \citep{lime}, Kernel Shapley \citep{lundberg2017unified}, and Sample Shapley \citep{strumbelj2010efficient}).
The results demonstrate that the post-hoc methods are stable, barely producing discrepant explanations under output probability perturbations.
The observation suggests that neural network models may be the primary source of fragile explanations.

\end{abstract}

\section{Introduction}
\label{se:introduction}
Despite the remarkable performance of neural network models in natural language processing (NLP), the lack of interpretability has raised much concern in terms of their reliability and trustworthiness \citep{zhang2021survey,doshi2017towards,hooker2019benchmark,jiang2018trust}. 
A common way to improve a model's interpretability is to generate explanations for its predictions from the post-hoc manner. 
We call these explanations post-hoc explanations \citep{doshi2017towards,molnar2018guide}.
Post-hoc explanations demonstrate the relationship between the input text and the model prediction by identifying feature importance scores \citep{du2019techniques}. 
In general, a feature with a higher importance score is more important in contributing to the prediction result. 
Based on feature importance scores, we can select top important features as the model explanation.

\begin{figure*}[htbp]
 \centering
 \includegraphics[width=0.8\textwidth]{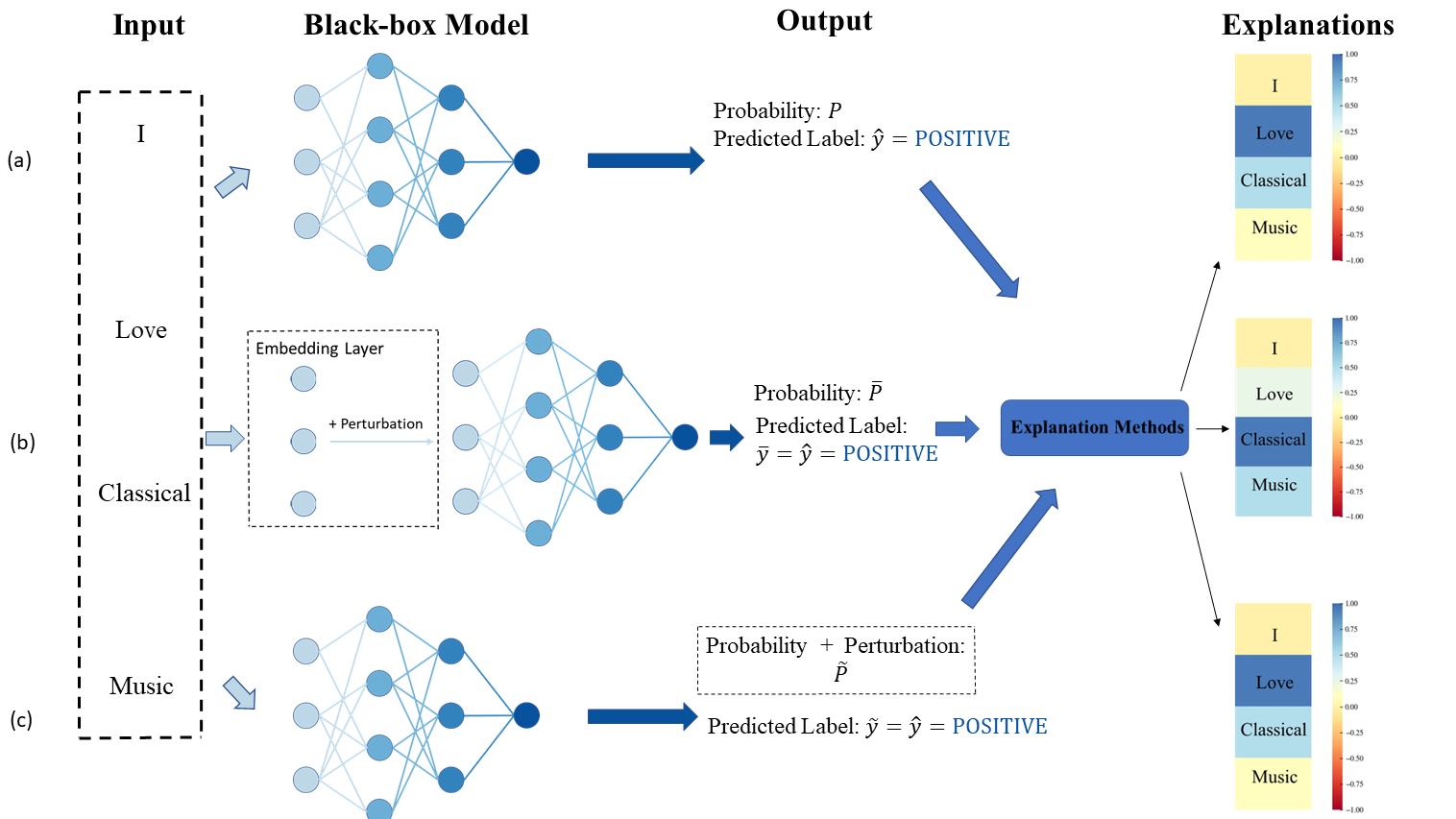}
 \caption{The pipeline of a simple example that post-hoc explanation methods generate explanations with (a) no perturbation applied. (b) perturbation applied at the input side. (c) perturbation applied at the output probabilities.}
 \label{fig:1}
\end{figure*} 

However, some recent works \citep{ghorbani2019interpretation,subramanya2019fooling,zhang2020interpretable,ivankay2022fooling, sinha2021perturbing} have observed explanation discrepancy when input-side perturbation is applied to the model. 
One question to this observation is what makes the explanation discrepant?
Explanations generated by a post-hoc method \citep{lime,lundberg2017unified,friedman2001greedy} depend on a model's prediction probabilities.
If perturbations at the input side cause model prediction probabilities to change, post-hoc explanations may change accordingly.

In \autoref{fig:1} (a), we demonstrate a simple example of the process that generates explanations using a post-hoc method. 
The explanation is generated depending on the probability $P$. 
In \autoref{fig:1} (b), we demonstrate an example of the same process with perturbation at the input side. 
The explanation is generated depending on the probability $\bar{P}$.
The output probabilities in the two examples are not the same, i.e. $P \ne \bar{P}$.
In \autoref{fig:1} (a) and (b), it is noticeable that the feature importance score of the same feature has changed. 
For instance, the feature ``love'' has different importance scores in the two examples.
Since feature importance scores are inconsistent, the explanations in the two examples are different. 
We call this \emph{explanation discrepancy}, which will be introduced more in \autoref{se:explanation_discrepancy}.

However, the prediction label in \autoref{fig:1} (a), $\hat{y}$, and the prediction label in \autoref{fig:1} (b), $\bar{y}$, are equal, which is $\hat{y} = \bar{y}=\textsc{Positive}$.
This indicates that input side perturbations may not flip the model prediction label, while can make output probabilities change, hence further leading to explanation discrepancy.
We argue that, under input side perturbations, it is difficult to identify the source causing the explanation discrepancy. 
One intuitive justification is that the perturbation at the input side has to pass through both the model and the post-hoc explanation method.
Both the model and the post-hoc explanation method are possible factors that result in unstable explanations.
For example, the model's prediction behavior may change under input side perturbations, that is focusing on different features to make predictions, hence resulting in the explanation discrepancy \citep{chen-ji-2020-learning, chen2022adversarial}. 
Or the explanation method itself may be vulnerable to input perturbations, producing discrepant explanations.
The instability may not be told from the prediction results, but reflected in the explanations, i.e., explanation discrepancy

In this paper, we propose a simple strategy to demonstrate the potential source that causes explanation discrepancy.
To circumvent the potential influence of the model on the explanations, we design an \emph{output probability perturbation} method by slightly modifying the prediction probabilities, as shown in \autoref{fig:1} (c).
In this work, we focus on the model-agnostic post-hoc methods, LIME \citep{lime}, Kernel Shapley \citep{lundberg2017unified}, and Sample Shapley \citep{strumbelj2010efficient}, that explain the black-box models. 
If a similar explanation discrepancy can be observed when only output probability perturbation is applied, it would suggest that post-hoc explanation methods may be unstable because the potential influence from the black-box model has been blocked.
Otherwise, we should not blame post-hoc explanation methods as the source of vulnerability in fragile explanations \citep{sinha2021perturbing,subramanya2019fooling}. 
\section{Method}
\subsection{Background}
\label{se:background}
For a text classification task, $\vec{x}$ denotes the input text consisting of $N$ words, $\vec{x}=[\vec{x}^{(1)}, \cdots, \vec{x}^{(N)}]$, with each component $\vec{x}^{(n)} \in {R}^{d}$ representing the $n$-th word embedding. 
We define a black-box classifier as $f(\cdot)$ and its output probability of a given $\vec{x}$ on the corresponding label $k$ is $P(y=k\mid\vec{x})=f_k(\vec{x})$, where $k\in\{1,\dots,C\}$ and $C$ is the total number of label classes.

To explain a black-box model's prediction $\hat{y} = f(\vec{x})$, a class of post-hoc explanation methods approximate the model locally via additive feature attributions \citep{NIPS2017_7062, lime, shrikumar2017learning}. 
Specifically, these algorithms demonstrate the relationship between the input text and the prediction result by evaluating the contribution of each input feature to the model prediction result. 
These methods would assign a feature importance score to each input feature to represent its contribution to the prediction. 
We use LIME \citep{lime} as an example. 

\paragraph{Example: Post-hoc Explanation Method, LIME.}
It first sub-samples words from the input, $\vec{x}$, to form a list of pseudo examples $\{\vec{z}_{j=1}^{L}\}$, and then the contributions of input features are estimated by a linear approximation $f_{\hat{y}}(\vec{r}) \approx g_{\hat{y}}(\vec{r'})$, where $\vec{r} \in \{\vec{x}, \vec{z}_{j=1}^{L}\}$, $g_{\hat{y}}(\vec{r})=\vec{w}_{\hat{y}}^{T}\vec{r'}$, and $\vec{r'}$ is a simple representation of $\vec{r}$, e.g. bag-of-words representation. 
The weights $\{w_{\hat{y}}^{(n)}\}$ represent importance scores of input features $\{\vec{x}^{(n)}\}$. Let $I(\vec{x},\hat{y},P)$ denote the explanation for the model prediction on $\vec{x}$, where $\hat{y}$ is the predicted label and $P$ represents output probabilities. 


\subsection{Explanation Discrepancy}
\label{se:explanation_discrepancy}
As mentioned in the previous section, the explanation discrepancy may happen when input perturbations are applied to the model.
Let $I(\vec{x},\hat{y},P)$ and $I(\bar{\vec{x}},\bar{y},\bar{P})$ denote the explanation to the model prediction based on the original input $\vec{x}$ and the perturbed input $\bar{\vec{x}}$ respectively, where $\bar{\vec{x}} = \vec{x} + \varepsilon$, and $\varepsilon$ is the perturbation at input. 
Similarly, we define $I(\vec{x},\widetilde{y},\widetilde{P})$ as the explanation to the prediction based on the perturbed output probability $\widetilde{P} = P + \varepsilon'$, where $\varepsilon'$ is the perturbation on the output probability. Note that when $\varepsilon$ and $\varepsilon'$ are small, the model prediction stay the same, which is $\hat{y} = \bar{y} = \widetilde{y}$.
The explanation discrepancy between $I(\bar{\vec{x}},\bar{y},\bar{P})$ and $I(\vec{x},\hat{y},P)$ is denoted as $\delta_{input}$, and the discrepancy between $I(\vec{x},\widetilde{y},\widetilde{P})$ and $I(\vec{x},\hat{y},P)$ is denoted as $\delta_{output}$.

We use \autoref{fig:1} in \autoref{se:introduction} as an example to illustrate explanation discrepancy in details. 
The explanation, $I(\vec{x},\hat{y},P)$, in \autoref{fig:1} (a) is ``Love", ``Classical", ``I" and ``Music", in the descending order of importance scores. 
The explanation, $I(\bar{\vec{x}},\bar{y},\bar{P})$, in \autoref{fig:1} (b) is ``Classical", ``Music", ``Love", and ``I", in the descending order. 
The explanation, $I(\vec{x},\hat{y},\widetilde{P})$, in \autoref{fig:1} (c) is ``Love", ``Classical", ``I" and ``Music", in the descending order. 
Generally, after perturbation, explanation inconsistency reflects in two aspects. 
The first aspect is whether the overall ranking of the features based on their importance scores in the explanation remains the same.  
For example, ``Love" ranks the first in the explanation in \autoref{fig:1} (a), while drops to the third in the explanation in \autoref{fig:1} (b).
The discrepancy is denoted as $\delta_{input}$.
The second aspect is whether the top $K$ important features in the explanation are consistent.
For example, if $K=2$, the first two important words in \autoref{fig:1} (a) are ``Love" and ``Classical", while those in \autoref{fig:1} (b) are ``Classical" and ``Music". 
The difference can also be denoted as $\delta_{input}$ mentioned above.
Similarly, the same aspect of explanation discrepancy in \autoref{fig:1} (a) and \autoref{fig:1} (c) can be denoted as $\delta_{output}$.

\subsection{Output Probability Perturbation Method}
As mentioned in \autoref{se:introduction}, the limitation of input perturbation methods is the difficulty in identifying the primary source that causes explanation discrepancy. 
Motivated by this, we propose the output probability perturbation method to circumvent the influence of black-box models.

Specifically, given an example $\vec{x}$, we add a small perturbation to the model output probabilities $\{P(y=k \mid \vec{x}) + \varepsilon'_{y=k}\}_{k=1}^{C}$.
To guarantee the modified $\{P(y=k \mid \vec{x}) + \varepsilon'_{y=k}\}_{k=1}^{C}$ are still legitimate probabilities, we further normalize them as 
\begin{equation}
    \widetilde{P}(y=k \mid \vec{x}) = \frac{P(y=k \mid \vec{x}) + \varepsilon'_{y=k}}{
        \sum_{i=1}^{C} \{P(y=i\mid \vec{x}) + \varepsilon'_{y=i}\}
    }
\end{equation}
The explanation in the case with output probability perturbation is computed based on the output probability $\widetilde{P}(y=\hat{y}\mid\vec{x})$. 
The proposed method well suits the motivation of investigating the source that causes explanation discrepancy. 
The main reason is that, unlike perturbation applied at the input side, the proposed method avoids the potential effects of the model's vulnerability on post-hoc explanations. 
We use LIME \citep{lime} as an example to demonstrate the proposed method.

\paragraph{Example: Output probability perturbation in LIME algorithm.}
As denoted in \autoref{se:background}, $\vec{r'}$ is the bag-of-words representation of the original input text, $\vec{x}$.
A simplified version \footnote{Without the example weight computed from a kernel function and the regularization term of explanation complexity.} of LIME algorithm is equivalent to finding a solution of the following linear equation:

\begin{equation}
\label{eq:linear}
     \vec{w}_{\hat{y}}^{T}\vec{r'} = \widetilde{\vec{p}}_{\hat{y}}
\end{equation}
where $\widetilde{\vec{p}}_{\hat{y}} = [\widetilde{P}(y=\hat{y}\mid\vec{x}),\widetilde{P}(y=\hat{y}\mid\vec{z}_1),\dots,\widetilde{P}(y=\hat{y}\mid\vec{z}_L)]^T$ are the perturbed probabilities on the label $\hat{y}$, and $\vec{w}_{\hat{y}}^{T}$ is the weight vector, where each element measures the contribution of an input word to the prediction $\hat{y}$. 
A typical explanation from LIME consists of top important words according to $\vec{w}_{\hat{y}}$.
Essentially, the proposed output perturbation is similar to the perturbation analysis in linear systems \citep{golub2013matrix}, which aims to identify the stability of these systems. 
Despite the simple formulation in Equation \ref{eq:linear}, a similar linear system can also be used to explain the Shapley-based explanation methods (e.g., Sample Shapley \citep{strumbelj2010efficient}).

\section{Experiment}
\subsection{Experiment Setup}
\label{se:experiment_setup}
\paragraph{Datasets.}
We adopt four text classification datasets: IMDB movie reviews dataset \citep[IMDB]{maas2011learning}, AG’s news dataset \citep[AG's News]{zhang2015character}, Stanford Sentiment Treebank dataset with binary labels \citep[SST-2]{socher2013recursive}, and 6-class questions classification dataset TREC \citep[TREC]{li2002learning}.
The summary statistics of datasets are shown in \autoref{tab:1}.

\begin{table*}[htbp]
\centering
\begin{tabular}{lllllllll}
\hline
Dataset & C & L & \textbf{$\#train$} & \textbf{$\#dev$} & \textbf{$\#test$} & vocab & threshold & length\\
\hline
IMDB & 2 & 268 & 20K & 5K & 25K & 29571 & 5 & 250 \\
SST-2 & 2 & 19 & 6920 & 872 & 1821 & 16190 & 0 & 50 \\
AG's News & 4 & 32 & 114K & 6K & 7.6K & 21838 & 5 & 50 \\
TREC & 6 & 10 & 5000 & 452 & 500 & 8026 & 0 & 15 \\
\hline
\end{tabular}
\caption{
Summary statistics for the datasets where C is the number of classes, L is the average sentence length, $\#$ counts the number of examples in train/dev/test sets, vocab is the vocab size, and the threshold is the low-frequency threshold, and length is mini-batch sentence length.}
\label{tab:1}
\end{table*}

\paragraph{Models.} 
We apply three neural network models, Convolutional Neural Network \citep[CNN]{kim-2014-convolutional}, Long Short Term Memory Network \citep[LSTM]{hochreiter1997long}, and Bidirectional Encoder Representations from Transformers \citep[BERT]{devlin2018bert}. 

The principle of CNN model is based on information processing in the visual system of humans. 
The core characteristics are that it can efficiently decrease the dimension of input, and it can efficiently retain important features of the input \citep{kim-2014-convolutional}. 

LSTM model is one advanced RNN model. 
Unlike the architecture of a standard feedforward deep learning neural network,  it has feedback connections in the architecture, which helps to process sequential data (e.g., language and speech) \citep[LSTM]{hochreiter1997long}. 

BERT model is a Language Model (LM). 
In the NLP research, the main tasks of the BERT model are (1) Sentence pairs classification tasks and (2) Single sentence classification tasks \citep{devlin2018bert}. 
In this work, we focus on the second task while we apply the BERT model in the experiment.

The prediction performance of the three models on the four datasets are recorded in \autoref{tab:2}.
\begin{table}[htbp]
\centering
\begin{tabular}{llll}
\hline
\textbf{Dataset} & \textbf{CNN} & \textbf{LSTM} & \textbf{BERT}\\
\hline
IMDB & 86.30 & 86.60 & 90.80\\
SST-2 & 82.48 & 80.83 & 91.82\\
AG's News & 89.90 & 88.90 & 95.10\\
TREC & 92.41 & 90.80 & 97.00\\
\hline
\end{tabular}
\caption{Prediction accuracy($\%$) of the three neural network models (CNN, LSTM and BERT) on the four datasets (IMDB, SST-2, AG's News and TREC).}
\label{tab:2}
\end{table}

\paragraph{Post-hoc Explanation Methods.}
We adopt three post-hoc explanation methods, Local Interpretable Model-Agnostic Explanations \citep[LIME]{lime}, Kernel Shapley \citep{lundberg2017unified}, and Sample Shapley \citep{strumbelj2010efficient}. 
LIME, Kernel Shapley, and Sample Shapley are additive feature attribution methods. 
The additive feature method provides a feature importance score on every feature for each text input based on the model prediction. 

LIME and Kernel Shapley are two post-hoc methods adopting a similar strategy. 
The first step is to generate a set of pseudo examples and their corresponding labels based on the black-box model's predictions on them \citep{lime,lundberg2017unified}.
The second step is to train an explainable machine learning model (eg: linear regression, LASSO) with the pseudo examples \citep{lime,lundberg2017unified}. 
The difference between the LIME algorithm and the Kernel Shapley algorithm is in the way to calculate the weight of pseudo examples in the explainable model \citep{molnar2018guide}.
LIME algorithm relies on the distance between the original example and the pseudo example \citep{lime}. 
Kernel Shapley algorithm relies on the Shapley value estimation \citep{lundberg2017unified}.

Sample Shapley is a post-hoc method based on Shapley value \citep{shapley1953value}, which stems from coalitional game theory. 
Shapley value provides an axiomatic solution to attribute the contribution of each word in a fair way. 
However, the exponential complexity of computing Shapley value is intractable. 
Sampling Shapley \citep{strumbelj2010efficient} provides a solvable approximation to Shapley value via sampling.

\paragraph{Evaluation Metrics.}
In the experiment, we apply two evaluation metrics, Kendall’s Tau order rank correlation score, and the Top-$K$ important words overlap score \citep{chen2019robust,kendall1938new,ghorbani2019interpretation} to evaluate the discrepancy between explanations (i.e., $\delta_{input}$ and $\delta_{output}$).

As illustrated in \autoref{se:explanation_discrepancy}, explanation discrepancy can be evaluated in in two aspects. 
We use Kendall’s Tau order rank correlation score to quantify the change of the overall ranking of feature importance scores in explanations. 
For example, in \autoref{fig:1} (a) and (b), we can apply Kendall’s Tau order rank correlation score to identify how close the overall ranking of features in the two examples.
If the score is close to $1$, then the two explanations are similar.  
If the score is close to $-1$, then the two explanations differ significantly. 
We use Top-$K$ important words overlap score to evaluate the discrepancy on the top $K$ features in the explanations. 
This metric computes the overlap ratio among the top $K$ features. 
In this work, we set $K=5$.
\subsection{Explanation Discrepancy Comparison Experiment}
\label{se:experiment_1}
To explore the primary source causing fragile explanations, we conduct a comparison experiment to evaluate and compare between explanation discrepancy $\delta_{input}$, and explanation discrepancy $\delta_{output}$. 
The definition of $\delta_{input}$, and $\delta_{output}$ are introduced in \autoref{se:explanation_discrepancy}. 
$\delta_{input}$ denotes the discrepancy between the explanation generated by the black-box model with no perturbation, $I(\vec{x},\hat{y},P)$, and the explanation generated by the black-box model with perturbation at the input, $I(\bar{\vec{x}},\hat{y},\bar{P})$. 
While $\delta_{output}$ denotes the discrepancy of $I(\vec{x},\hat{y},P)$ and the explanation generated by the black-box model with perturbation at the output probability, $I(\bar{\vec{x}},\hat{y},\widetilde{P})$.

In this experiment, for output probability perturbation, we directly add random noise to the model output probabilities. 
For comparison, we add the noise to word embeddings for input perturbations \citep{liu2020adversarial}.
Both input side perturbation and output probability perturbation are applied with noise sampled from a Gaussian distribution, $\mathcal{N}(0,\sigma^2)$.
We apply Gaussian noise because it is easy to control the perturbation level by modifying the variance of the Gaussian distribution $\sigma^2$.  
In experiments, we applied five different perturbation levels from ``$0$'' to ``$4$''. 
``$0$'' means the slightest perturbation level, zero perturbation, while ``$4$'' represents the strongest perturbation level. 
The specific value of each perturbation level is shown in \autoref{tab:3}. 
Note that for each level, the input side perturbations and the output probability perturbations are different because we select different perturbations for the input side and the output probability to reach a similar accuracy at each level. 
If the model's accuracy is not close at each level, it is difficult to evaluate the results. 
\begin{table}[htbp]
\centering
\begin{tabular}{lll}
\hline
 \textbf{Perturbation Source} & \textbf{Level}& $\sigma^{2}$ \\
\hline
Input Side ($\sigma^{2}_{input}$)& 0 & 0 \\
      & 1 & 0.05 \\
      & 2 & 0.1 \\
      & 3 & 0.15\\
      & 4 & 0.2 \\
\hline
Output Probability ($\sigma^{2}_{output}$)& 0 & 0  \\
      & 1 & 0.25\\
      & 2 & 0.5 \\
      & 3 & 0.75 \\
      & 4 & 1 \\
\hline
\end{tabular}
\caption{Perturbation levels applied to the input and output respectively.}
\label{tab:3}
\end{table}
\subsection{Results and Discussion}
\label{se:results}

\begin{figure*}[htbp]
        \centering
        \subfigure[]{\includegraphics[width=0.3\linewidth]{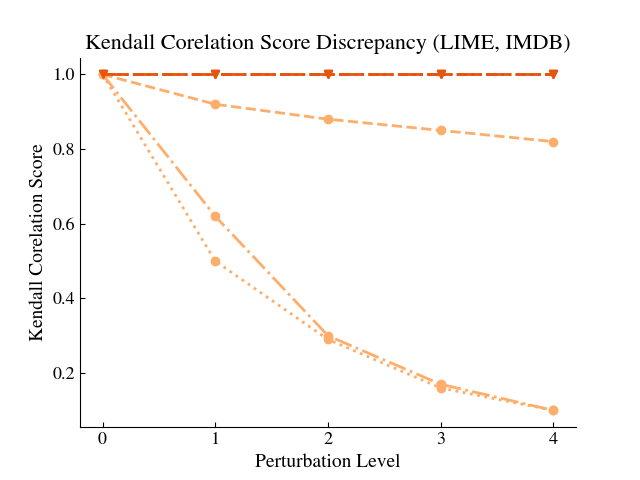}}
        \subfigure[]{\includegraphics[width=0.3\linewidth]{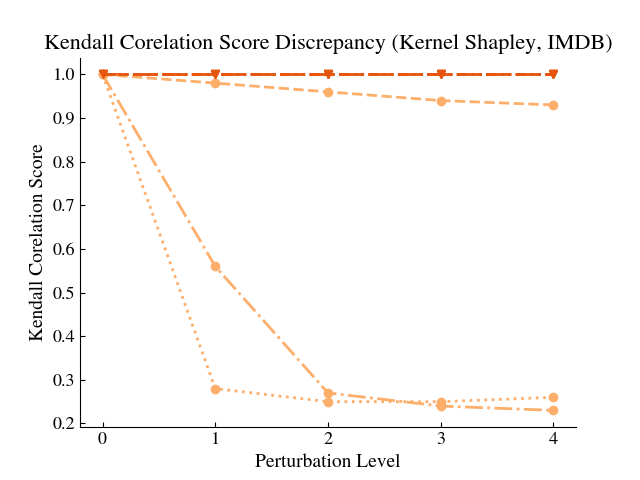}}
        \subfigure[]{\includegraphics[width=0.3\textwidth]{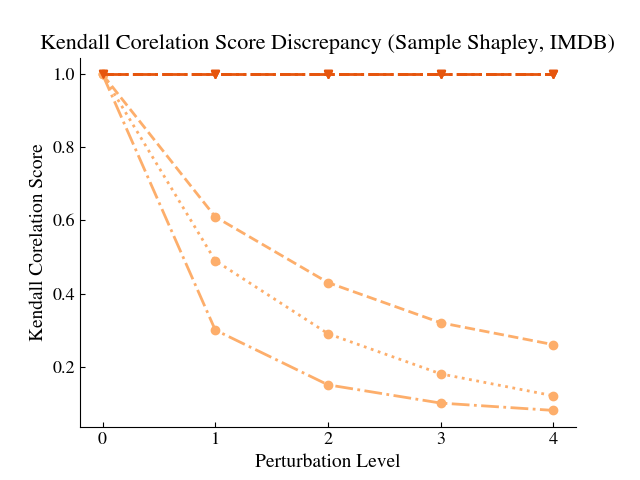}}
        \\
        \centering
        \subfigure[]{\includegraphics[width=0.3\linewidth]{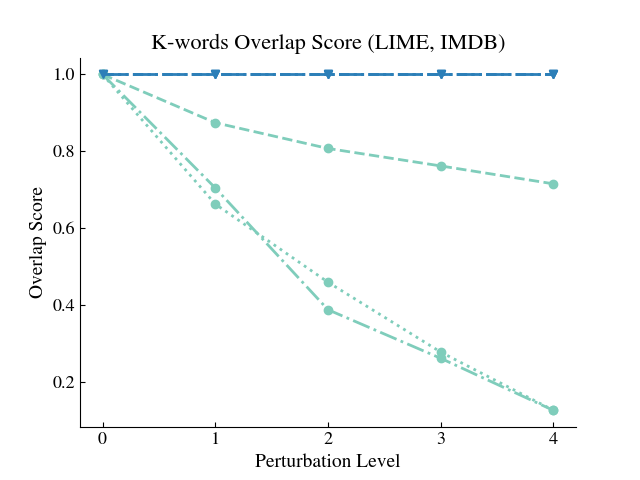}}
        \subfigure[]{\includegraphics[width=0.3\linewidth]{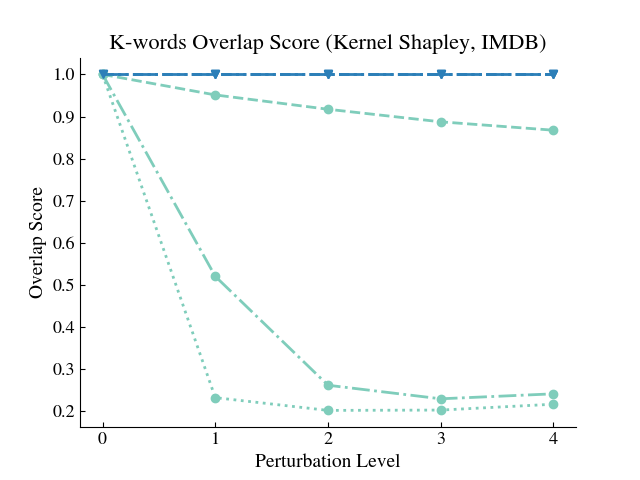}}
        \subfigure[]{\includegraphics[width=0.3\textwidth]{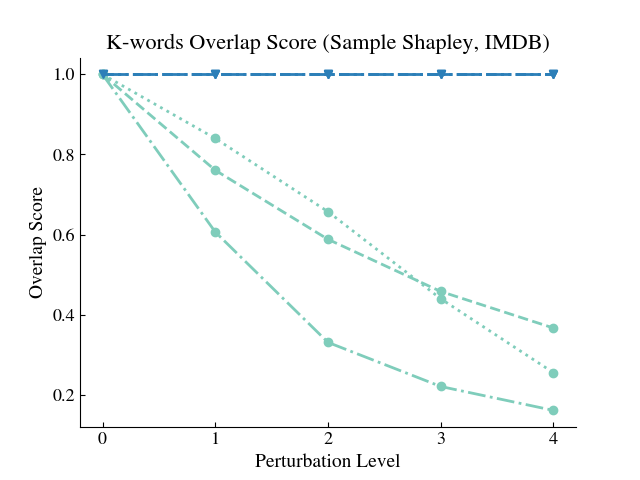}}
        \\
        \centering
        \subfigure{\includegraphics[width=0.8\textwidth]{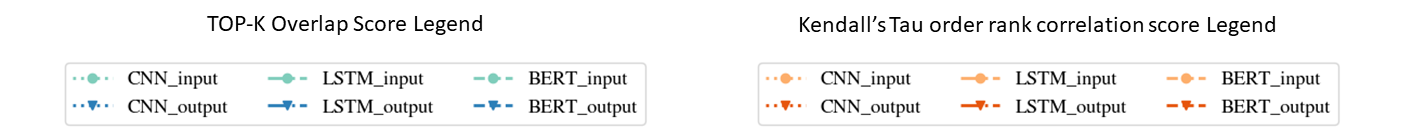}}
        \caption{Comparison experiment results on the IMDB dataset; (a) and (d) demonstrate results using LIME method; (b) and (e) demonstrate results using Kernel Sharply method; (c) and (f) demonstrate results using Sample Sharply method.}
        \label{fig:3.1}
    \end{figure*}

\autoref{fig:3.1} shows the results of the IMDB dataset. 
Due to the page limit, full results of other datasets are shown in \autoref{fig:3.2}, \autoref{fig:3.3} and \autoref{fig:3.4} in \autoref{sec:appendix_b}, which have similar tendencies.
Kendall's Tau order rank correlation score plots are shown in \autoref{fig:3.1} (a), (b) and (c). 
Top-$K$ important words overlap score plots are shown in \autoref{fig:3.1} (d), (e) and (f). 
\autoref{fig:3.1} (a) and (d) show the results of the LIME method.
\autoref{fig:3.1} (b) and (e) show the results of the Kernel Shapley method.
\autoref{fig:3.1} (c) and (f) show the results of the Sample Shapley method.
\paragraph{Kendall's Tau order rank correlation score evaluation results.}
Kendall's Tau order rank correlation score results indirectly illustrate the stability of post-hoc explanation methods.
Furthermore, previous observation on the explanation difference can be attributed to the potential influence caused by the black-box model.   
In \autoref{fig:3.1} (a), (b) and (c), the large gap between $\delta_{input}$ and $\delta_{output}$ is consistent across all three post-hoc explanation methods, LIME, Kernel Shapley and Sample Shapley. 
For output probability perturbations,  it is noticeable that the values of Kendall's Tau order rank correlation scores remain the same with the perturbation level increasing from ``0'' to ``4".
This indicates that the overall ranking of feature importance scores are stable under output perturbations. 
Furthermore, the results suggest that for a given input, if $\vec{x}$ and prediction results stay unchanged, $\hat{y} = \widetilde{y}$, the only perturbation $\varepsilon'$ at output probability is unlikely to influence explanations generated by the post-hoc methods. 
In other words, the explanation discrepancy observed in the previous study \citep{ivankay2022fooling, sinha2021perturbing} is unlikely caused by the post-hoc methods.
Meanwhile, for the baseline results (perturbation applied at the input), it is notifiable that the values of Kendall's Tau order rank correlation scores decrease obviously with the increase of input perturbation intensity levels.
This indicates that the black-box model is vulnerable to input perturbations, which causes fragile explanations. 
Based on the observations, we claim that the black-box model is more likely to be the primary source that results in fragile explanations.
\paragraph{Top-$K$ word importance score evaluation results.}
Top-$K$ word importance score evaluation shows the same result: the black-box model is the primary source causing explanation discrepancy. 
In \autoref{fig:3.1} (d), (e) and (f), $\delta_{input}$ against $\delta_{output}$ display an obvious discrepancy across the three post-hoc explanation methods. 
For output probability perturbations, $\delta_{output}$ shows no change in the overlap among the top $K$ important words. 
This indicates that, for the top five important features in the explanation of each corresponding prediction result, output probability perturbations will not cause any difference.
The results under this metric also illustrate that the black-box model is more likely to cause fragile explanations than explanation methods themselves.

\subsection{Further Analysis on LIME Algorithm}
\label{se:analysis}
According to the previous results, we have a conclusion that post-hoc explanation methods are stable. 
We further analyze the stability of the explanation algorithms. 
We use the LIME algorithm \citep{lime} as an example. 

\begin{equation}
\label{eq:loss_function}
     L(f_{\hat{y}}(\vec{r}),g_{\hat{y}}(\vec{r'}),\pi) = \sum_{\vec{r},\vec{r'}\in{\vec{R}}} \pi(f_{\hat{y}}(\vec{r}) - g_{\hat{y}}(\vec{r'}))
\end{equation}

\autoref{eq:loss_function} is definition of the loss function in LIME algorithm \citep{lime}. 
In the loss function, $\pi g_{\hat{y}}(\vec{r'})$ is denoted the kernel calculation function of the algorithm.
$\vec{r'}$ represents the pseudo example based on the original example, $\vec{r}$. 
$g_{\hat{y}}(\vec{r'})$ represents the linear local explainable model on the pseudo example, $\vec{r'}$. 
Here, we use a general linear model representation to represent the explainable model, $g_{\hat{y}}(\vec{r})=\vec{w}_{\hat{y}}^{T}\vec{r'}$.
$\vec{w}_{\hat{y}}^{T}$ is the weight function of the linear model and it is the importance feature score calculation function as well. 
\autoref{eq:kernel} is the kernel calculation function of the LIME algorithm after expanding. 

\begin{equation}
\label{eq:kernel}
     G = \pi g_{\hat{y}}(\vec{r})
       =\pi \vec{w}_{\hat{y}}^{T}\vec{r'}
\end{equation}

The form of the kernel calculation function can be interpreted as a general linear function, $Ax=b$.
In the linear function, the condition number, $(\kappa)$, is applied to evaluate how sensitive the linear function is due to a small change at the input and reflects in its output \citep{belsley2005regression}.
If the condition number, $(\kappa)$, which is the largest eigenvalue in the matrix $A$ divided by the smallest eigenvalue in the matrix $A$, is large, the solution $x$ would change rapidly by a slight difference in $b$, which would cause sensitivity of the solution to the slight error in the input \citep{Goodfellow-et-al-2016}. 
In \autoref{eq:kernel}, $\pi \vec{r'}$ can be considered as the matrix $A$, and the feature importance score function $\vec{w}_{\hat{y}}^{T}$ can be considered as the solution $x$. 
If $\pi \vec{r'}$ is a stable linear system,  the feature importance score function $\vec{w}_{\hat{y}}^{T}$ would be unlikely sensitive to a minor change at the linear system input side, and the corresponding post-hoc explanation method is stable.  
The form of the kernel calculation function can be interpreted as a general linear function, $Ax=b$.
In the linear function, the condition number, $(\kappa)$, is applied to evaluate how sensitive the linear function is due to a small change at the input and reflects in its output \citep{belsley2005regression}.
If the condition number, $(\kappa)$, which is the largest eigenvalue in the matrix $A$ divided by the smallest eigenvalue in the matrix $A$, is large, the solution $x$ would change rapidly by a slight difference in $b$, which would cause sensitivity of the solution to the slight error in the input \citep{Goodfellow-et-al-2016}. 
In \autoref{eq:kernel}, $\pi \vec{r'}$ can be considered as the matrix $A$, and the feature importance score function $vec{w}_{\hat{y}}^{T}$ can be considered as the solution $\vec{r}$. 
If $\pi \vec{r'}$ is a stable linear system,  the feature importance score function $\vec{w}_{\hat{y}}^{T}$ would be unlikely sensitive to a minor change at the linear system input side, and the corresponding post-hoc explanation method is stable.   
Since the kernel function is a pure numerical step without semantics involved. 
We conduct a simulation experiment to explore the stability of the LIME algorithm \citep{lime}.
\paragraph{Simulation Experiment and Results} 
In the simulation experiment, the pseudo example, $\vec{r'}$, is a matrix with the size of sub-sampling size, $m$, multiple with the length of a sentence, $l$. 
We select $m = 200$, which is the actual sample size value we applied in the comparison experiment. 
For the sentence length, first, we simulate the case when sentence length is fixed, that is $l=20$. Then, to compare condition number distribution when sentence length is different,  we apply two more cases, that are $l = 30$,  and $l = 40$. 
For each length, we simulate it for $500$ iterations.
For $\pi$, it is the distance between the original input to the sub-sampling based on the original input in the LIME algorithm \citep{lime}. In the simulation experiment, we apply cosine distance to represent the value of $\pi$. 

\begin{table}[htbp]
\centering
\begin{tabular}{lll}
\hline
\textbf{Total iteration number} & \textbf{$\kappa\in[5,6)$} & \textbf{$\kappa\in[6,7)$}\\
\hline
500 & 392 & 108\\
\hline
\end{tabular}
\caption{Condition number $(\kappa)$ distribution when $l=20$}
\label{tab:7}
\end{table}

\begin{figure}[htbp]
 \centering
 \includegraphics[width=0.5\textwidth]{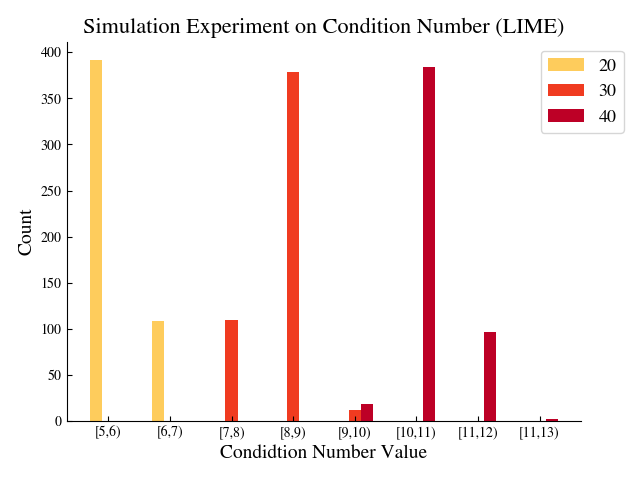}
 \caption{Simulation Experiment Result.}
 \label{fig:3.5}
\end{figure} 

In \autoref{tab:7}, the result of the simulation experiment when the sentence length is fixed, $l = 20$, demonstrates that the majority of the condition number $\kappa$ of the matrix $\pi \vec{r'}$ is lower than $7$. 
In Goldstein's work, it suggests that the condition number of a stable or a well-conditioning matrix should be lower than $30$ \citep{goldstein1993conditioning}.
It means that the feature importance score function, $W$, is less likely influenced by a small perturbation involved, which is also reflected in the real dataset results in the comparison experiment in \autoref{se:results}. 
In \autoref{fig:3.5}, the result of the simulation experiment when sentence lengths are different shows that when the length of the sentence increases, the condition number $\kappa$ of the matrix $\pi \vec{r'}$ increases with a tiny amplitude. 
The majority of the condition number $\kappa$ of the matrix is lower than $13$ when the length is from $20$ to $40$. 
Although the result demonstrates that the condition number $\kappa$ would increase with sentence length increasing,  the increasing amplitude is small and the majority of the condition number is lower than the threshold number. 
The result suggests that the matrix $\pi \vec{r'}$ in the LIME algorithm can remain a small condition number, which makes the linear system relatively stable. 
In other words, the LIME algorithm \citep{lime} is a relatively stable post-hoc explanation method. 

\section{Previous Works}
\paragraph{Post-hoc Explanation Methods} Most works focus on explaining neural network models in a post-hoc manner, especially generating a local explanation for each model prediction. 
The white-box explanation methods, such as gradient-based explanations \citep{hechtlinger2016interpretation}, and attention-based explanations \citep{ghaeini2018interpreting}, either require additional information (e.g. gradients) from the model or incur much debates regarding their faithfulness to model predictions \citep{jain2019attention}. 

Another line of work focuses on explaining black-box models in the model-agnostic way. 
\citet{li2016understanding} proposed a perturbation-based explanation method, Leave-one-out, that attributes feature importance to model predictions by erasing input features one by one. 
\citet{lime} proposed to estimate feature contributions locally via linear approximation based on pseudo examples. 
Some other works proposed the variants of the Shapley value \citep{shapley1953quota} to measure feature importance, such as the Sample Shapley method \citep{strumbelj2010efficient}, the Kernel Shapley method \citep{lundberg2017unified}, and the L/C-Shapley method\citep{chen2018shapley}.
\paragraph{Model robustness} Recent works have shown the vulnerability of model due to adversarial attacks \citep{szegedy2013intriguing, goodfellow2014explaining, zhao2017generating}. 
Some adversarial examples are similar to original examples but can quickly flip model predictions, which causes concern on model robustness \citep{jia2019certified}. 
In the text domain, a common way to generate adversarial examples is by heuristically manipulating the input text, such as replacing words with their synonyms \citep{alzantot2018generating, ren2019generating, jin2020bert}, misspelling words \citep{li2018textbugger, gao2018black}, inserting/removing words \citep{liang2017deep}, or concatenating triggers \citep{wallace2019universal}. 
\paragraph{Explanation Robustness} Previous work explored explanation robustness by either perturbing the inputs \citep{ghorbani2019interpretation, subramanya2019fooling, zhang2020interpretable, heo2019fooling} or manipulating the model \citep{wang2020gradient, slack2020fooling, zafar2021lack}. 
For example, Slack's group fooled post-hoc explanation methods by hiding the bias for black-box models based on the proposed novel scaffolding technique \citep{slack2020fooling}. 
However, all of these works cannot disentangle the sources that cause fragile explanation. 
Differently, the proposed method mitigates the influence of model to the explanation by perturbing model outputs.
\label{se:previous_works}

\section{Conclusion}
In this work, our main contribution is to identify the primary source of fragile explanation, where we propose an output probability perturbation method. 
With the help of this proposed method, observation results can illustrate a conclusion that the primary potential source that caused fragile explanations in the previous studies is the black-box model itself, which also illustrate that some limitations of prior methods.
Furthermore, in \autoref{se:analysis}, we analyze the kernel calculation inside the LIME algorithm \citep{lime}. 
We apply the condition number of the matrix and simulation experiments to demonstrate that the kernel calculation matrix inside LIME has a low condition number. 
This result further suggests the stability of the LIME algorithm. 

\clearpage
\bibliography{anthology,custom}

\begin{thebibliography}{55}
\expandafter\ifx\csname natexlab\endcsname\relax\def\natexlab#1{#1}\fi

\bibitem[{Alzantot et~al.(2018)Alzantot, Sharma, Elgohary, Ho, Srivastava, and
  Chang}]{alzantot2018generating}
Moustafa Alzantot, Yash Sharma, Ahmed Elgohary, Bo-Jhang Ho, Mani Srivastava,
  and Kai-Wei Chang. 2018.
\newblock Generating natural language adversarial examples.
\newblock \emph{arXiv preprint arXiv:1804.07998}.

\bibitem[{Belsley et~al.(2005)Belsley, Kuh, and Welsch}]{belsley2005regression}
David~A Belsley, Edwin Kuh, and Roy~E Welsch. 2005.
\newblock \emph{Regression diagnostics: Identifying influential data and
  sources of collinearity}.
\newblock John Wiley \& Sons.

\bibitem[{Chen and Ji(2020)}]{chen-ji-2020-learning}
Hanjie Chen and Yangfeng Ji. 2020.
\newblock \href {https://doi.org/10.18653/v1/2020.emnlp-main.347} {Learning
  variational word masks to improve the interpretability of neural text
  classifiers}.
\newblock In \emph{Proceedings of the 2020 Conference on Empirical Methods in
  Natural Language Processing (EMNLP)}, pages 4236--4251, Online. Association
  for Computational Linguistics.

\bibitem[{Chen and Ji(2022)}]{chen2022adversarial}
Hanjie Chen and Yangfeng Ji. 2022.
\newblock Adversarial training for improving model robustness? look at both
  prediction and interpretation.
\newblock In \emph{Proceedings of the AAAI Conference on Artificial
  Intelligence}.

\bibitem[{Chen et~al.(2018)Chen, Song, Wainwright, and
  Jordan}]{chen2018shapley}
Jianbo Chen, Le~Song, Martin~J Wainwright, and Michael~I Jordan. 2018.
\newblock L-shapley and c-shapley: Efficient model interpretation for
  structured data.
\newblock \emph{arXiv preprint arXiv:1808.02610}.

\bibitem[{Chen et~al.(2019)Chen, Wu, Rastogi, Liang, and Jha}]{chen2019robust}
Jiefeng Chen, Xi~Wu, Vaibhav Rastogi, Yingyu Liang, and Somesh Jha. 2019.
\newblock Robust attribution regularization.
\newblock \emph{arXiv preprint arXiv:1905.09957}.

\bibitem[{Devlin et~al.(2018)Devlin, Chang, Lee, and
  Toutanova}]{devlin2018bert}
Jacob Devlin, Ming-Wei Chang, Kenton Lee, and Kristina Toutanova. 2018.
\newblock Bert: Pre-training of deep bidirectional transformers for language
  understanding.
\newblock \emph{arXiv preprint arXiv:1810.04805}.

\bibitem[{Doshi-Velez and Kim(2017)}]{doshi2017towards}
Finale Doshi-Velez and Been Kim. 2017.
\newblock Towards a rigorous science of interpretable machine learning.
\newblock \emph{arXiv preprint arXiv:1702.08608}.

\bibitem[{Du et~al.(2019)Du, Liu, and Hu}]{du2019techniques}
Mengnan Du, Ninghao Liu, and Xia Hu. 2019.
\newblock Techniques for interpretable machine learning.
\newblock \emph{Communications of the ACM}, 63(1):68--77.

\bibitem[{Friedman(2001)}]{friedman2001greedy}
Jerome~H Friedman. 2001.
\newblock Greedy function approximation: a gradient boosting machine.
\newblock \emph{Annals of statistics}, pages 1189--1232.

\bibitem[{Gao et~al.(2018)Gao, Lanchantin, Soffa, and Qi}]{gao2018black}
Ji~Gao, Jack Lanchantin, Mary~Lou Soffa, and Yanjun Qi. 2018.
\newblock Black-box generation of adversarial text sequences to evade deep
  learning classifiers.
\newblock In \emph{2018 IEEE Security and Privacy Workshops (SPW)}, pages
  50--56. IEEE.

\bibitem[{Ghaeini et~al.(2018)Ghaeini, Fern, and
  Tadepalli}]{ghaeini2018interpreting}
Reza Ghaeini, Xiaoli~Z Fern, and Prasad Tadepalli. 2018.
\newblock Interpreting recurrent and attention-based neural models: a case
  study on natural language inference.
\newblock \emph{arXiv preprint arXiv:1808.03894}.

\bibitem[{Ghorbani et~al.(2019)Ghorbani, Abid, and
  Zou}]{ghorbani2019interpretation}
Amirata Ghorbani, Abubakar Abid, and James Zou. 2019.
\newblock Interpretation of neural networks is fragile.
\newblock In \emph{Proceedings of the AAAI Conference on Artificial
  Intelligence}, volume~33, pages 3681--3688.

\bibitem[{Goldstein(1993)}]{goldstein1993conditioning}
Richard Goldstein. 1993.
\newblock Conditioning diagnostics: Collinearity and weak data in regression.

\bibitem[{Golub and Van~Loan(2013)}]{golub2013matrix}
Gene~H Golub and Charles~F Van~Loan. 2013.
\newblock \emph{Matrix computations}, volume~3.
\newblock JHU press.

\bibitem[{Goodfellow et~al.(2016)Goodfellow, Bengio, and
  Courville}]{Goodfellow-et-al-2016}
Ian Goodfellow, Yoshua Bengio, and Aaron Courville. 2016.
\newblock \emph{Deep Learning}.
\newblock MIT Press.
\newblock \url{http://www.deeplearningbook.org}.

\bibitem[{Goodfellow et~al.(2014)Goodfellow, Shlens, and
  Szegedy}]{goodfellow2014explaining}
Ian~J Goodfellow, Jonathon Shlens, and Christian Szegedy. 2014.
\newblock Explaining and harnessing adversarial examples.
\newblock \emph{arXiv preprint arXiv:1412.6572}.

\bibitem[{Hechtlinger(2016)}]{hechtlinger2016interpretation}
Yotam Hechtlinger. 2016.
\newblock Interpretation of prediction models using the input gradient.
\newblock \emph{arXiv preprint arXiv:1611.07634}.

\bibitem[{Heo et~al.(2019)Heo, Joo, and Moon}]{heo2019fooling}
Juyeon Heo, Sunghwan Joo, and Taesup Moon. 2019.
\newblock Fooling neural network interpretations via adversarial model
  manipulation.
\newblock \emph{arXiv preprint arXiv:1902.02041}.

\bibitem[{Hochreiter and Schmidhuber(1997)}]{hochreiter1997long}
Sepp Hochreiter and J{\"u}rgen Schmidhuber. 1997.
\newblock Long short-term memory.
\newblock \emph{Neural computation}, 9(8):1735--1780.

\bibitem[{Hooker et~al.(2019)Hooker, Erhan, Kindermans, and
  Kim}]{hooker2019benchmark}
Sara Hooker, Dumitru Erhan, Pieter-Jan Kindermans, and Been Kim. 2019.
\newblock A benchmark for interpretability methods in deep neural networks.
\newblock \emph{Advances in neural information processing systems}, 32.

\bibitem[{Ivankay et~al.(2022)Ivankay, Girardi, Marchiori, and
  Frossard}]{ivankay2022fooling}
Adam Ivankay, Ivan Girardi, Chiara Marchiori, and Pascal Frossard. 2022.
\newblock Fooling explanations in text classifiers.
\newblock \emph{arXiv preprint arXiv:2206.03178}.

\bibitem[{Jain and Wallace(2019)}]{jain2019attention}
Sarthak Jain and Byron~C Wallace. 2019.
\newblock Attention is not explanation.
\newblock \emph{arXiv preprint arXiv:1902.10186}.

\bibitem[{Jia et~al.(2019)Jia, Raghunathan, G{\"o}ksel, and
  Liang}]{jia2019certified}
Robin Jia, Aditi Raghunathan, Kerem G{\"o}ksel, and Percy Liang. 2019.
\newblock Certified robustness to adversarial word substitutions.
\newblock \emph{arXiv preprint arXiv:1909.00986}.

\bibitem[{Jiang et~al.(2018)Jiang, Kim, Guan, and Gupta}]{jiang2018trust}
Heinrich Jiang, Been Kim, Melody Guan, and Maya Gupta. 2018.
\newblock To trust or not to trust a classifier.
\newblock \emph{Advances in neural information processing systems}, 31.

\bibitem[{Jin et~al.(2020)Jin, Jin, Zhou, and Szolovits}]{jin2020bert}
Di~Jin, Zhijing Jin, Joey~Tianyi Zhou, and Peter Szolovits. 2020.
\newblock Is bert really robust? a strong baseline for natural language attack
  on text classification and entailment.
\newblock In \emph{Proceedings of the AAAI conference on artificial
  intelligence}, volume~34, pages 8018--8025.

\bibitem[{Kendall(1938)}]{kendall1938new}
Maurice~G Kendall. 1938.
\newblock A new measure of rank correlation.
\newblock \emph{Biometrika}, 30(1/2):81--93.

\bibitem[{Kim(2014)}]{kim-2014-convolutional}
Yoon Kim. 2014.
\newblock \href {https://doi.org/10.3115/v1/D14-1181} {Convolutional neural
  networks for sentence classification}.
\newblock In \emph{Proceedings of the 2014 Conference on Empirical Methods in
  Natural Language Processing ({EMNLP})}, pages 1746--1751, Doha, Qatar.
  Association for Computational Linguistics.

\bibitem[{Li et~al.(2018)Li, Ji, Du, Li, and Wang}]{li2018textbugger}
Jinfeng Li, Shouling Ji, Tianyu Du, Bo~Li, and Ting Wang. 2018.
\newblock Textbugger: Generating adversarial text against real-world
  applications.
\newblock \emph{arXiv preprint arXiv:1812.05271}.

\bibitem[{Li et~al.(2016)Li, Monroe, and Jurafsky}]{li2016understanding}
Jiwei Li, Will Monroe, and Dan Jurafsky. 2016.
\newblock Understanding neural networks through representation erasure.
\newblock \emph{arXiv preprint arXiv:1612.08220}.

\bibitem[{Li and Roth(2002)}]{li2002learning}
Xin Li and Dan Roth. 2002.
\newblock Learning question classifiers.
\newblock In \emph{COLING 2002: The 19th International Conference on
  Computational Linguistics}.

\bibitem[{Liang et~al.(2017)Liang, Li, Su, Bian, Li, and Shi}]{liang2017deep}
Bin Liang, Hongcheng Li, Miaoqiang Su, Pan Bian, Xirong Li, and Wenchang Shi.
  2017.
\newblock Deep text classification can be fooled.
\newblock \emph{arXiv preprint arXiv:1704.08006}.

\bibitem[{Liu et~al.(2020)Liu, Cheng, He, Chen, Wang, Poon, and
  Gao}]{liu2020adversarial}
Xiaodong Liu, Hao Cheng, Pengcheng He, Weizhu Chen, Yu~Wang, Hoifung Poon, and
  Jianfeng Gao. 2020.
\newblock Adversarial training for large neural language models.
\newblock \emph{arXiv preprint arXiv:2004.08994}.

\bibitem[{Lundberg and Lee(2017{\natexlab{a}})}]{lundberg2017unified}
Scott~M Lundberg and Su-In Lee. 2017{\natexlab{a}}.
\newblock A unified approach to interpreting model predictions.
\newblock In \emph{Proceedings of the 31st international conference on neural
  information processing systems}, pages 4768--4777.

\bibitem[{Lundberg and Lee(2017{\natexlab{b}})}]{NIPS2017_7062}
Scott~M Lundberg and Su-In Lee. 2017{\natexlab{b}}.
\newblock \href
  {http://papers.nips.cc/paper/7062-a-unified-approach-to-interpreting-model-predictions.pdf}
  {A unified approach to interpreting model predictions}.
\newblock In I.~Guyon, U.~V. Luxburg, S.~Bengio, H.~Wallach, R.~Fergus,
  S.~Vishwanathan, and R.~Garnett, editors, \emph{Advances in Neural
  Information Processing Systems 30}, pages 4765--4774. Curran Associates, Inc.

\bibitem[{Maas et~al.(2011)Maas, Daly, Pham, Huang, Ng, and
  Potts}]{maas2011learning}
Andrew Maas, Raymond~E Daly, Peter~T Pham, Dan Huang, Andrew~Y Ng, and
  Christopher Potts. 2011.
\newblock Learning word vectors for sentiment analysis.
\newblock In \emph{Proceedings of the 49th annual meeting of the association
  for computational linguistics: Human language technologies}, pages 142--150.

\bibitem[{Molnar(2018)}]{molnar2018guide}
Christoph Molnar. 2018.
\newblock A guide for making black box models explainable.
\newblock \emph{URL: https://christophm. github. io/interpretable-ml-book}.

\bibitem[{Ren et~al.(2019)Ren, Deng, He, and Che}]{ren2019generating}
Shuhuai Ren, Yihe Deng, Kun He, and Wanxiang Che. 2019.
\newblock Generating natural language adversarial examples through probability
  weighted word saliency.
\newblock In \emph{Proceedings of the 57th annual meeting of the association
  for computational linguistics}, pages 1085--1097.

\bibitem[{Ribeiro et~al.(2016)Ribeiro, Singh, and Guestrin}]{lime}
Marco~Tulio Ribeiro, Sameer Singh, and Carlos Guestrin. 2016.
\newblock "why should {I} trust you?": Explaining the predictions of any
  classifier.
\newblock In \emph{Proceedings of the 22nd {ACM} {SIGKDD} International
  Conference on Knowledge Discovery and Data Mining, San Francisco, CA, USA,
  August 13-17, 2016}, pages 1135--1144.

\bibitem[{Shapley(1953{\natexlab{a}})}]{shapley1953value}
Lloyd~S Shapley. 1953{\natexlab{a}}.
\newblock A value for n-person games.
\newblock \emph{Contributions to the Theory of Games}, 2(28).

\bibitem[{Shapley(1953{\natexlab{b}})}]{shapley1953quota}
LS~Shapley. 1953{\natexlab{b}}.
\newblock Quota solutions op n-person games1.
\newblock \emph{Edited by Emil Artin and Marston Morse}, page 343.

\bibitem[{Shrikumar et~al.(2017)Shrikumar, Greenside, and
  Kundaje}]{shrikumar2017learning}
Avanti Shrikumar, Peyton Greenside, and Anshul Kundaje. 2017.
\newblock Learning important features through propagating activation
  differences.
\newblock In \emph{International Conference on Machine Learning}, pages
  3145--3153. PMLR.

\bibitem[{Sinha et~al.(2021)Sinha, Chen, Sekhon, Ji, and
  Qi}]{sinha2021perturbing}
Sanchit Sinha, Hanjie Chen, Arshdeep Sekhon, Yangfeng Ji, and Yanjun Qi. 2021.
\newblock Perturbing inputs for fragile interpretations in deep natural
  language processing.
\newblock \emph{arXiv preprint arXiv:2108.04990}.

\bibitem[{Slack et~al.(2020)Slack, Hilgard, Jia, Singh, and
  Lakkaraju}]{slack2020fooling}
Dylan Slack, Sophie Hilgard, Emily Jia, Sameer Singh, and Himabindu Lakkaraju.
  2020.
\newblock Fooling lime and shap: Adversarial attacks on post hoc explanation
  methods.
\newblock In \emph{Proceedings of the AAAI/ACM Conference on AI, Ethics, and
  Society}, pages 180--186.

\bibitem[{Socher et~al.(2013)Socher, Perelygin, Wu, Chuang, Manning, Ng, and
  Potts}]{socher2013recursive}
Richard Socher, Alex Perelygin, Jean Wu, Jason Chuang, Christopher~D Manning,
  Andrew~Y Ng, and Christopher Potts. 2013.
\newblock Recursive deep models for semantic compositionality over a sentiment
  treebank.
\newblock In \emph{Proceedings of the 2013 conference on empirical methods in
  natural language processing}, pages 1631--1642.

\bibitem[{Strumbelj and Kononenko(2010)}]{strumbelj2010efficient}
Erik Strumbelj and Igor Kononenko. 2010.
\newblock An efficient explanation of individual classifications using game
  theory.
\newblock \emph{The Journal of Machine Learning Research}, 11:1--18.

\bibitem[{Subramanya et~al.(2019)Subramanya, Pillai, and
  Pirsiavash}]{subramanya2019fooling}
Akshayvarun Subramanya, Vipin Pillai, and Hamed Pirsiavash. 2019.
\newblock Fooling network interpretation in image classification.
\newblock In \emph{Proceedings of the IEEE/CVF International Conference on
  Computer Vision}, pages 2020--2029.

\bibitem[{Szegedy et~al.(2013)Szegedy, Zaremba, Sutskever, Bruna, Erhan,
  Goodfellow, and Fergus}]{szegedy2013intriguing}
Christian Szegedy, Wojciech Zaremba, Ilya Sutskever, Joan Bruna, Dumitru Erhan,
  Ian Goodfellow, and Rob Fergus. 2013.
\newblock Intriguing properties of neural networks.
\newblock \emph{arXiv preprint arXiv:1312.6199}.

\bibitem[{Wallace et~al.(2019)Wallace, Feng, Kandpal, Gardner, and
  Singh}]{wallace2019universal}
Eric Wallace, Shi Feng, Nikhil Kandpal, Matt Gardner, and Sameer Singh. 2019.
\newblock Universal adversarial triggers for attacking and analyzing nlp.
\newblock \emph{arXiv preprint arXiv:1908.07125}.

\bibitem[{Wang et~al.(2020)Wang, Tuyls, Wallace, and Singh}]{wang2020gradient}
Junlin Wang, Jens Tuyls, Eric Wallace, and Sameer Singh. 2020.
\newblock Gradient-based analysis of nlp models is manipulable.
\newblock \emph{arXiv preprint arXiv:2010.05419}.

\bibitem[{Zafar et~al.(2021)Zafar, Donini, Slack, Archambeau, Das, and
  Kenthapadi}]{zafar2021lack}
Muhammad~Bilal Zafar, Michele Donini, Dylan Slack, C{\'e}dric Archambeau,
  Sanjiv Das, and Krishnaram Kenthapadi. 2021.
\newblock On the lack of robust interpretability of neural text classifiers.
\newblock \emph{arXiv preprint arXiv:2106.04631}.

\bibitem[{Zhang et~al.(2015)Zhang, Zhao, and LeCun}]{zhang2015character}
Xiang Zhang, Junbo Zhao, and Yann LeCun. 2015.
\newblock Character-level convolutional networks for text classification.
\newblock \emph{arXiv preprint arXiv:1509.01626}.

\bibitem[{Zhang et~al.(2020)Zhang, Wang, Shen, Ji, Luo, and
  Wang}]{zhang2020interpretable}
Xinyang Zhang, Ningfei Wang, Hua Shen, Shouling Ji, Xiapu Luo, and Ting Wang.
  2020.
\newblock Interpretable deep learning under fire.
\newblock In \emph{29th $\{$USENIX$\}$ Security Symposium ($\{$USENIX$\}$
  Security 20)}.

\bibitem[{Zhang et~al.(2021)Zhang, Ti{\v{n}}o, Leonardis, and
  Tang}]{zhang2021survey}
Yu~Zhang, Peter Ti{\v{n}}o, Ale{\v{s}} Leonardis, and Ke~Tang. 2021.
\newblock A survey on neural network interpretability.
\newblock \emph{IEEE Transactions on Emerging Topics in Computational
  Intelligence}.

\bibitem[{Zhao et~al.(2017)Zhao, Dua, and Singh}]{zhao2017generating}
Zhengli Zhao, Dheeru Dua, and Sameer Singh. 2017.
\newblock Generating natural adversarial examples.
\newblock \emph{arXiv preprint arXiv:1710.11342}.

\end{thebibliography}

\clearpage

\appendix

\section{Figures of Comparison Experiments Result on SST-2, AG's News and TREC Dataset}
\label{sec:appendix_b}
\begin{figure*}[htbp]
        \centering
        \subfigure[]{\includegraphics[width=0.3\linewidth]{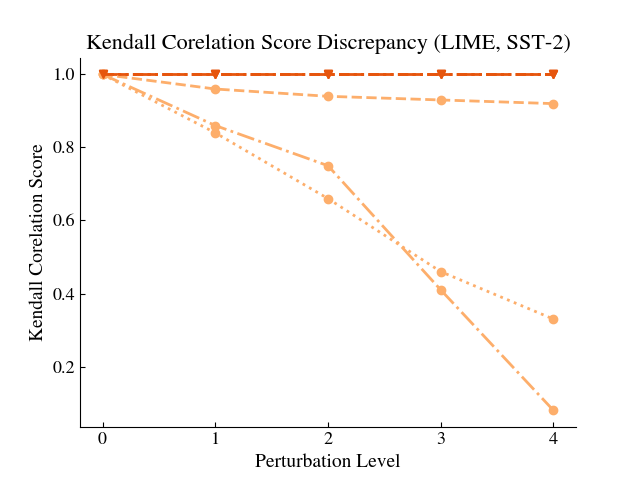}}
        \subfigure[]{\includegraphics[width=0.3\linewidth]{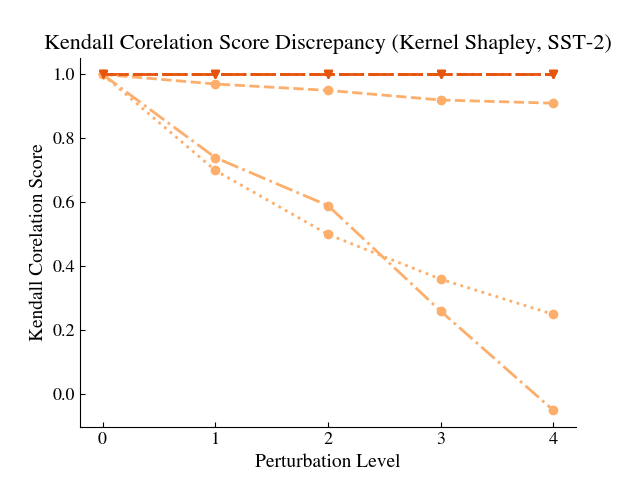}}
        \subfigure[]{\includegraphics[width=0.3\linewidth]{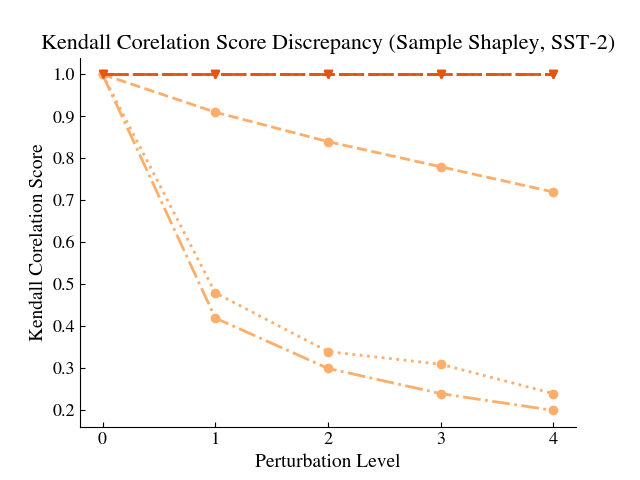}}
        \\
        \centering
        \subfigure[]{\includegraphics[width=0.3\linewidth]{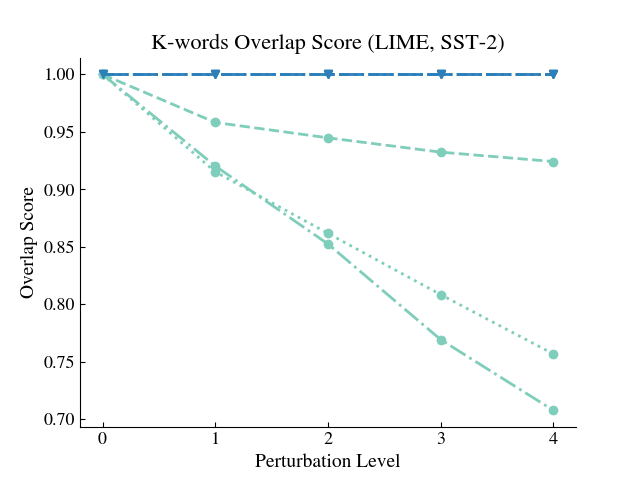}}
        \subfigure[]{\includegraphics[width=0.3\linewidth]{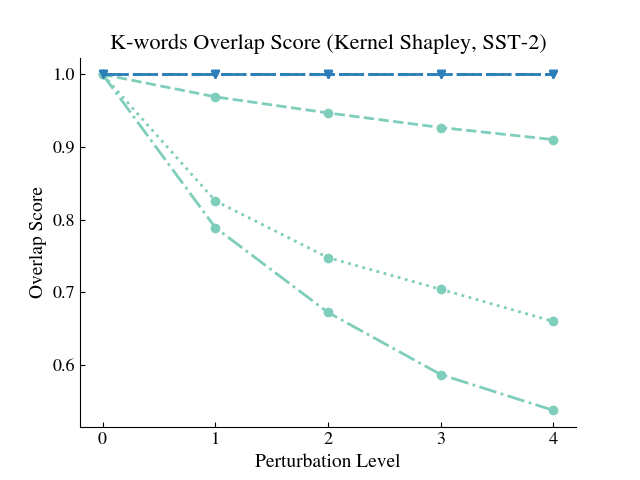}}
        \subfigure[]{\includegraphics[width=0.3\linewidth]{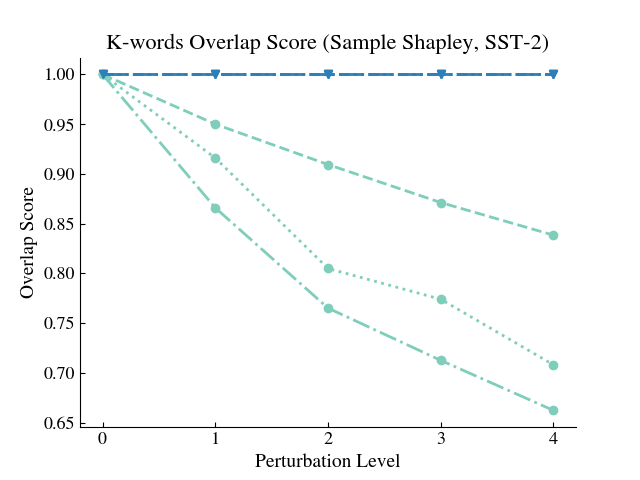}}
        \\
        \centering
        \subfigure{\includegraphics[width=0.8\textwidth]{plots/leng.png}}
        \caption{Comparison experiment results on the SST-2 dataset; (a) and (d) demonstrate results using LIME method; (b) and (e) demonstrate results using Kernel Sharply method; (c) and (f) demonstrate results using Sample Sharply method.}
        \label{fig:3.2}
    \end{figure*}

\begin{figure*}[htbp]
        \centering
        \subfigure[]{\includegraphics[width=0.3\linewidth]{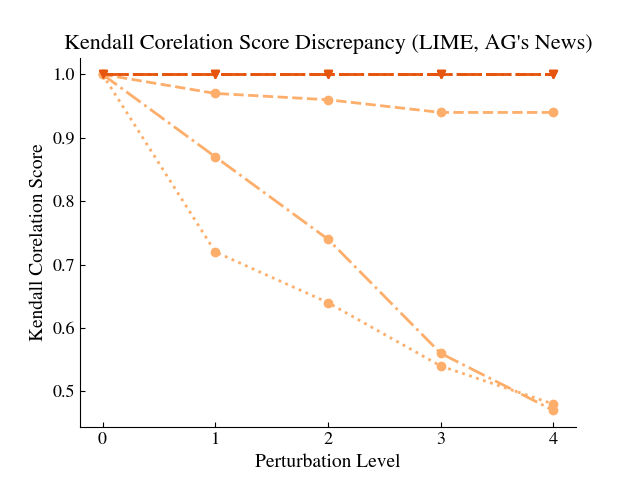}}
        \subfigure[]{\includegraphics[width=0.3\linewidth]{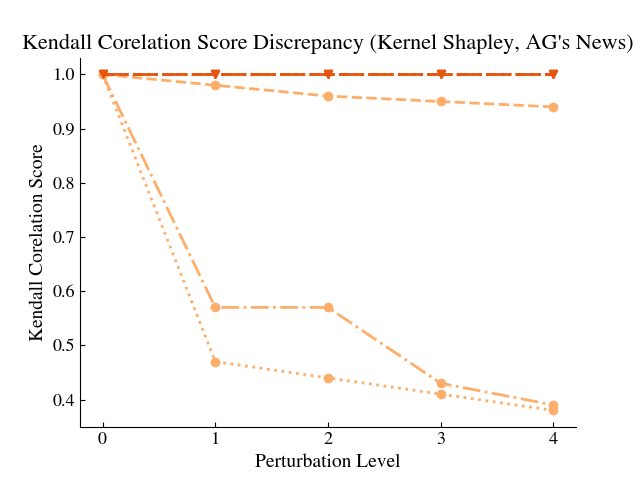}}
        \subfigure[]{\includegraphics[width=0.3\linewidth]{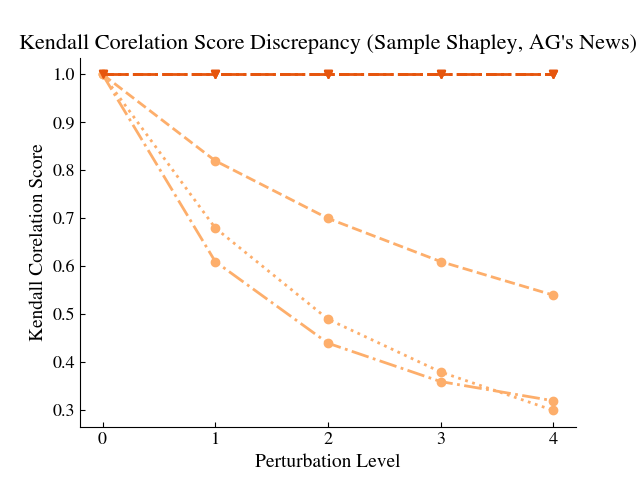}}
        \\
        \centering
        \subfigure[]{\includegraphics[width=0.3\linewidth]{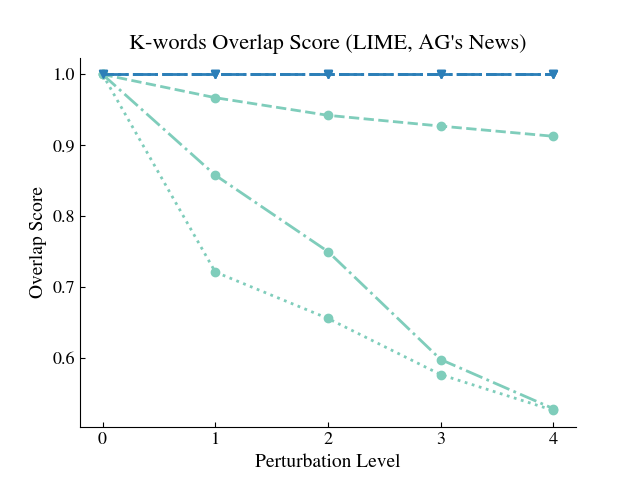}}
        \subfigure[]{\includegraphics[width=0.3\linewidth]{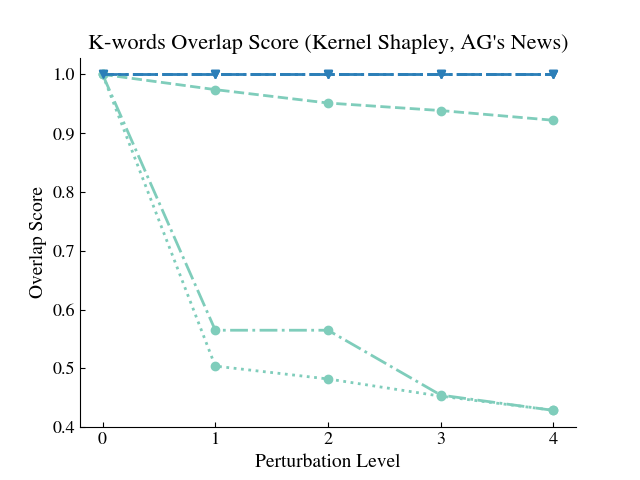}}
        \subfigure[]{\includegraphics[width=0.3\linewidth]{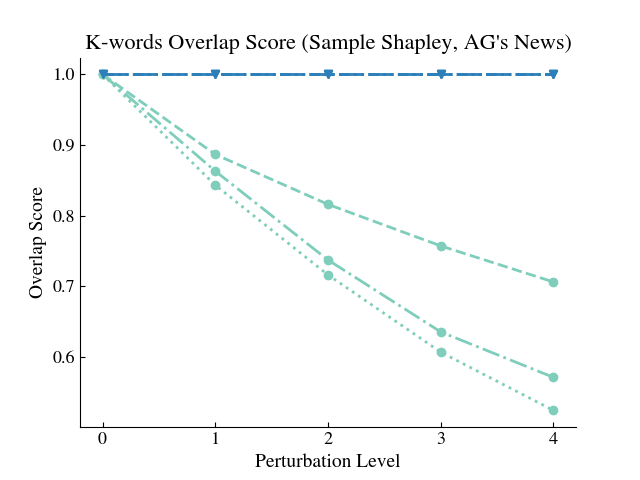}}
        \\
        \centering
        \subfigure{\includegraphics[width=0.8\textwidth]{plots/leng.png}}
        \caption{Comparison experiment results on the AG's News dataset; (a) and (d) demonstrate results using LIME method; (b) and (e) demonstrate results using Kernel Sharply method; (c) and (f) demonstrate results using Sample Sharply method.}
        \label{fig:3.3}
    \end{figure*}

\begin{figure*}[htbp]
        \centering
        \subfigure[]{\includegraphics[width=0.3\linewidth]{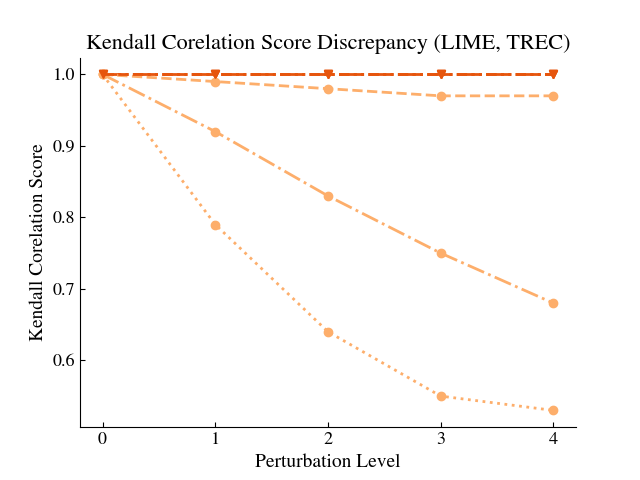}}
        \subfigure[]{\includegraphics[width=0.3\linewidth]{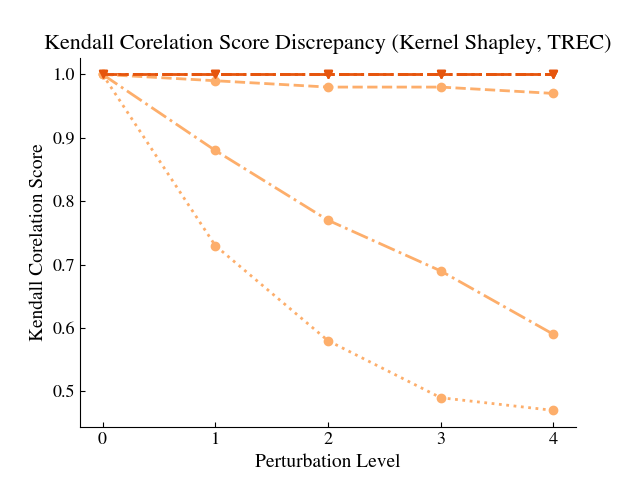}}
        \subfigure[]{\includegraphics[width=0.3\linewidth]{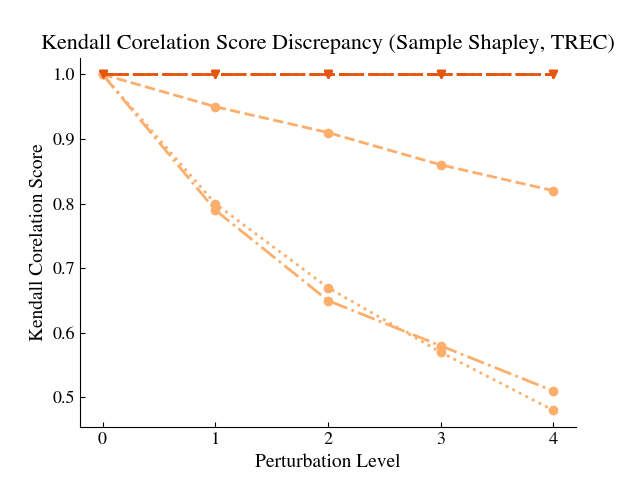}}
        \\
        \centering
        \subfigure[]{\includegraphics[width=0.3\linewidth]{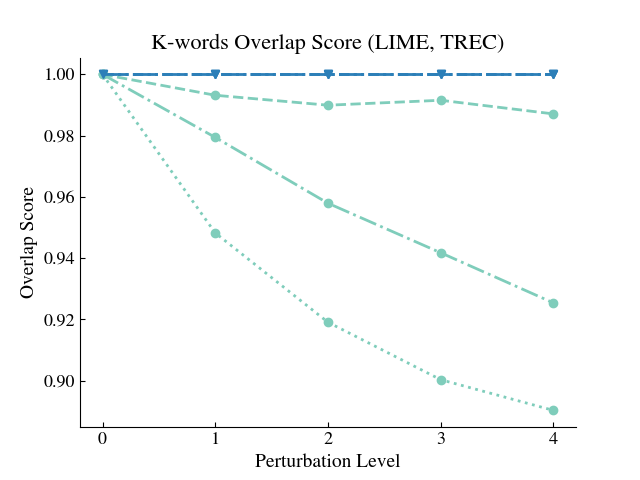}}
        \subfigure[]{\includegraphics[width=0.3\linewidth]{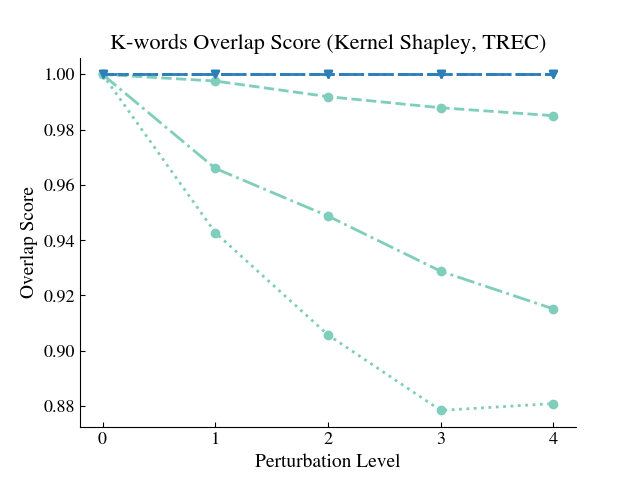}}
        \subfigure[]{\includegraphics[width=0.3\linewidth]{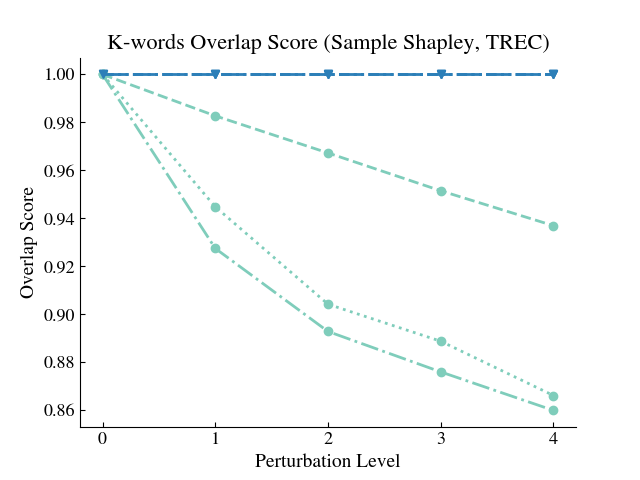}}
        \\
        \centering
        \subfigure{\includegraphics[width=0.8\textwidth]{plots/leng.png}}
        \caption{Comparison experiment results on the TREC dataset; (a) and (d) demonstrate results using LIME method; (b) and (e) demonstrate results using Kernel Sharply method; (c) and (f) demonstrate results using Sample Sharply method.}
        \label{fig:3.4}
    \end{figure*}

In the section, we include comparison experiments results of the SST-2 dataset in \autoref{fig:3.2}, the AG's News dataset in \autoref{fig:3.3}, and the TREC dataset in \autoref{fig:3.4}.


\end{document}